\documentclass{article}

     \usepackage[final]{neurips_2022}

\usepackage[utf8]{inputenc} 
\usepackage[T1]{fontenc}    
\usepackage[hidelinks]{hyperref}       
\usepackage{url}            
\usepackage{booktabs}       
\usepackage{amsfonts}       
\usepackage{nicefrac}       
\usepackage{microtype}      
\usepackage{xcolor}         

\usepackage{graphicx} 
\usepackage{algorithm}
\usepackage{algpseudocode}
\usepackage{amsmath,amssymb}
\usepackage{amsthm}
\usepackage{todonotes}
\usepackage[font=small,skip=0pt]{caption}
\usepackage[symbol]{footmisc}

\title{Target-independent XLA optimization using Reinforcement Learning}

\author{%
  Milan Ganai\thanks{Work conducted during an internship at Amazon.} \\
  University of California San Diego\\
  \texttt{mganai@ucsd.edu} \\
  \And
  Haichen Li \\
  Amazon \\
  \texttt{lhaiche@amazon.com} \\
  \And
  Theodore Enns \\
  Amazon \\
  \texttt{ennst@amazon.com} \\
  \And
  Yida Wang \\
  Amazon \\
  \texttt{wangyida@amazon.com} \\
  \And
  Randy Huang \\
  Amazon \\
  \texttt{renfu@amazon.com} \\
}

\begin{document}

\maketitle

\begin{abstract}
An important challenge in Machine Learning compilers like XLA is multi-pass optimization and analysis. There has been recent interest chiefly in XLA target-dependent optimization on the graph-level, subgraph-level, and kernel-level phases. We specifically focus on target-independent optimization XLA HLO pass ordering: our approach aims at finding the optimal sequence of compiler optimization passes, which is decoupled from target-dependent optimization. However, there is little domain specific study in pass ordering for XLA HLO. To this end, we propose introducing deep Reinforcement Learning (RL) based search for optimal XLA HLO pass ordering. We also propose enhancements to the deep RL algorithms to further improve optimal search performance and open the research direction for domain-specific guidance for RL. We create an XLA Gym experimentation framework as a tool to enable RL algorithms to interact with the compiler for passing optimizations and thereby train agents. Overall, in our experimentation we observe an average of $13.3\%$ improvement in operation count reduction on a benchmark of GPT-2 training graphs and $10.4\%$ improvement on a diverse benchmark including GPT-2, BERT, and ResNet graphs using the proposed approach over the compiler's default phase ordering.

\end{abstract}

\section{Introduction}
Machine Learning frameworks use Machine Learning compilers to convert neural networks into hardware readable specific code. They primarily use heuristic based approaches to solve optimizations problems at the different levels of the compiler stack. In the past several years, search-based machine learning methodologies have been proposed to optimize on the various levels of the compiler stack. Approaches have looked into sub-graph and kernel (fused operation nodes) level optimizations, optimizations of specific compiler passes, or joint optimizations across the graph, sub-graph, and kernel levels. However, the problem of graph level (for instance the High Level Operations (HLO) Intermediate Representation (IR) in XLA~\cite{frostig2018compiling}) pass ordering has been largely optimized by heuristic based approaches.

We explore multi-pass compiler optimization on Machine Learning compilers on the graph-level. We specifically aim to algebraically optimize Machine Learning XLA graphs with target-independent HLO compiler optimization passes. That is, we need to select the sequence of XLA HLO compiler optimization passes to transform the graph to optimize a specific objective. This objective may be instruction count, XLA operation count, or graph-size. Our contributions to this problem domain is as follows: 1) we introduce deep Reinforcement Learning algorithms to the problem space of XLA target-independent pass ordering optimizations, 2) we propose domain-specific enhancements to the deep RL algorithms in order to further improve their performance, and 3) we demonstrate the efficacy of our approaches in comparison with default. 

To enable RL algorithms to interact with the compiler and train the agents, we convert the problem into a Markov Decision process framework by creating an XLA Gym infrastructure based on OpenAI's Gym, described in Section~\ref{section:Gym}. We subsequently test various deep Reinforcement Learning algorithms in Section~\ref{section:EvaluateRL}, and we propose and test our enhancements in deep Reinforcement Learning algorithms in XLA Gym in Section~\ref{section:EvaluateEnhancement}.

\section{Related Works}

\subsection{LLVM phase ordering}
In LLVM~\cite{Lattner2004llvm}, a closely related problem is phase ordering~\cite{Ashouri2018survey, Chen2010Evaluating} which is the problem of selecting and ordering LLVM compiler optimizations. Various techniques have been proposed in phase ordering including collaborative filtering~\cite{Cereda2020}, design space exploration~\cite{Nobre2016}, and Bayesian Networks~\cite{Ashouri2016}. Usage of (deep) Reinforcement Learning has been seen in various LLVM compiler optimization problems such as PolyGym~\cite{Brauckmann2021} for polyhedral loop transformations, NeuroVectorizer~\cite{HajAli2020NeuroVectorizer} for single step instruction vectorization, and MLGPO~\cite{Trofin2021} for inlining for size.  RL-based approaches for phase ordering have been explored in Autophase~\cite{HajAli2020Autophase} and CORL~\cite{Mammadli2020}. We refer the reader to~\cite{MLinCompiler} for a more in depth survey of AI based techniques for LLVM compiler optimizations.

There are also several compiler optimization research tools such as OpenTuner~\cite{Ansel2014}, and YaCoS~\cite{Zanella2020} which are autotuning frameworks and ComPy-Learn~\cite{Brauckmann2020} for program representation. A notable environment is CompilerGym~\cite{Cummins2022} which consists of various compiler optimization problems presented using the OpenAI Gym interface including LLVM phase ordering. Overall, literature on the compiler optimization pass ordering approaches have largely focused on the LLVM environments, but Machine Learning compiler target-independent pass ordering decoupled from target-dependent optimization has not been as extensively explored.

\subsection{ML compiler frameworks}
Within the domain of ML compilers, there has been a growth in research in optimizations across the various levels of the compiler stack. Datasets such as Tenset~\cite{Zheng2021} for tensor compilers have been produced for offline learning. Autotuning has been proposed in the subgraph and kernel levels in works such as TVM~\cite{Chen2018TVM}, AutoTVM~\cite{Chen2018OptimizeTensor}, Ansor~\cite{Zheng2020Ansor}, FlexTensor~\cite{Zheng2020Flextensor}, Halide~\cite{Adams2019}, Chameleon~\cite{Ahn2020}, AdaTune~\cite{Li2020Adatune}, and Tensor Comprehension~\cite{Vasilache2018}. Some of these approaches are currently being utilized in production-level compilers such as the one in the AWS Neuron SDK~\footnote[2]{AWS Neuron SDK documentation site: \url{https://awsdocs-neuron.readthedocs-hosted.com/}}~\cite{awsneurondoc2022}. Operator-level optimizations and code generation for custom hardware accelerators has been explored in AKG~\cite{Zhao2020} and Mind Mappings~\cite{Hegde2021}. Our methodology operates on target-independent HLO graph level in determining the optimal pass ordering — graph/sub-graph and kernel level autotuning as well as operator-level optimizations for hardware-specific optimization is orthogonal to our work. The Value Learning approach of~\cite{Steiner2021} does full graph loop optimization and is effective for a single compiler stage. Reinforcement Learning based computational graph optimization (GO)~\cite{Zhou2020} jointly optimizes device placement, operator fusion, and operator scheduling does not focus on multiple target-independent passes. A survey of Deep Learning compilers can be found in~\cite{DLCompiler}. Overall, we introduce Reinforcement Learning to multiple compiler pass optimization and demonstrate improvement on a target-independent level.

\section{Preliminaries}

\subsection{XLA}
The proposal for a Reinforcement Learning framework may generalize to any Machine Learning compiler framework. In this paper, we test in XLA specifically~\cite{TensorFlowXLA}. XLA is a Machine Learning compiler that generates code for a variety of hardware targets. The compilation process can be divided into a graph level phase, kernel level hardware lowering phase, and a low level target specific phase. In the first phase XLA uses a High Level Operation Intermediate Representation which is a graph of the tensor computations. In this target-independent step, various compiler optimization and analysis passes transform the graph into an algebraically optimized output HLO graph. The nodes composing this graph are fused operations and are known as kernels. These are brought down to the target to be converted into instructions specific to the hardware. In the process, the graph is converted into various IRs like Loop IR and Backend IR. Finally, in the hardware phase, low level target specific optimizations are are performed on the machine instructions.

\subsection{Markov Decision Processes}
We formulate the problem of XLA multi-pass optimization as a Markov Decision Process, which can be represented by the 5-tuple $\langle\mathcal{S},\mathcal{A},T,R,\gamma\rangle$. Specifically, $\mathcal{S}$ represents the state space, which is the set of all possible values of the observable features of the graph at a given point in time. $\mathcal{A}$ is the action space that consists of 53 HLO compiler optimization passes that can be used to optimize a given graph at each step. $T:\mathcal{S}\times\mathcal{A}\times\mathcal{S}\rightarrow[0,1]$ is the transition function which signifies the probability $T(s'|s,a)$ that an agent at state $s$ taking some action $a$ will transition to state $s'$. This is in essence an abstraction of the compiler taking a graph with some features $s$ and an optimization pass $a$ and outputting a new optimized graph with features $s'$. $R:\mathcal{S}\rightarrow \mathbb{R}$ is a reward function that determines the immediate gain $r(s)$ of an agent being in a particular state $s$. Finally, the discount factor $0<<\gamma<1$ discourages postponing good actions. A policy $\pi_\theta:\mathcal{S}\times\mathcal{A}\rightarrow[0,1]$ parameterized by $\theta$ provides the probability $\pi_\theta(a|s)$ that an agent will take action $a$ given it is in state $s$. The goal is to learn the optimal policy $\pi^*$ that maximizes expected cumulative discounted reward, i.e. returns.

\section{Gym Functionality}
\label{section:Gym}
For simulating the Markov Decision Process for the XLA multi-pass optimization, we use the OpenAI Gym~\cite{OpenAI} structure and create the XLA Gym environments. Specifically, this requires defining and engineering the following:

\indent{\em State:} The state space indicates the set of values the observable features of the state of the graph can take. Because there is a wide range of values that XLA Gym operation count types can take, this is primarily represented as an array indicating minimum and maximum values of each feature.

\indent{\em Action:} Similarly, the action space defines the set of values that represent actions. In pass ordering, it is most feasible to define the action space as discrete by mapping each action to a distinct number from 0 to one less than the total number of actions.

\indent{\em Reward:} The reward must be manually engineered in order to best optimize for guiding the reinforcement learning agent. Because we are looking to reduce the overall operation count and for ease of calculation, the reward is a function simply of the observation features. However, if we want to discourage certain types of actions or transitions, the reward function can easily generalize to become a function of the transition (i.e. $R(s,a,s')$).

\indent{\em Info:} An information dictionary is returned at each step of Gym. This provides environment designers to conveniently provide any additional information if needed such as additional cost information.

\indent{\em Reset:} Beginning each learning episode, it is important to reset the environment in order to bring the agent back to a starting state and clear any needed environment variables before proceeding.

\indent{\em Step:} Once an action has been chosen by the RL agent, the environment acts as a classic decision chain by taking the action and returning the next state and reward. Furthermore, gym environments provide a boolean indicating termination of an episode and the information dictionary.

\indent{\em Render:} Rendering allows for a readable/parsable output at any given instance of the environment.

\subsection{XLA Gym environment}
\begin{figure}[!hbt]
\centering
\includegraphics[width=\textwidth]{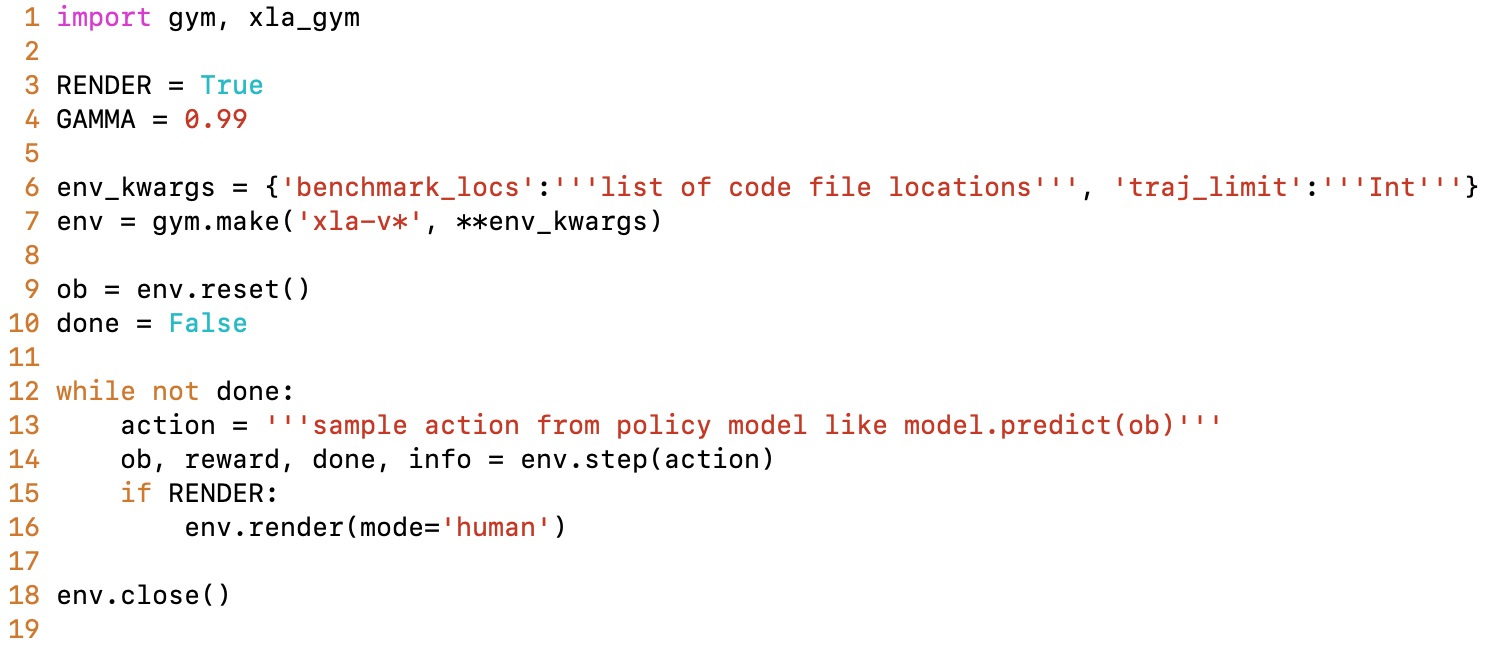}
\caption{Example usage of XLA Gym's standard environment.}
\label{fig:CodeXLA}
\end{figure}

Using the OpenAI Gym structure, we develop XLA Gym environments suitable for various use cases. The environments are built off of the basic environment. In general, the states are arrays with elements indicating various types of XLA operation count, and the actions are whole numbers less than the total number of 53 HLO compiler optimization passes. The reward is negative of the scaled XLA operation count, though may vary across the environments. Resetting initializes a new graph without optimizations and returns the initial observable state. At each step, an action is provided, and the next state, reward, boolean for episodic termination, and info dictionary are returned. Depending on the environment, the info dictionary may contain HLO cost analysis features such as FLOP count and transcendental count. Figure~\ref{fig:CodeXLA} shows the basic usage of the XLA Gym environment.

\section{Experiments}
\subsection{Evaluating Deep RL Algorithms}
\label{section:EvaluateRL}
With our XLA Gym structure, we proceed to explore various deep Reinforcement Learning Algorithms. Deep RL algorithms contain off-policy algorithms, which train a policy different from the one used to generate and collect environment data, and on-policy algorithms, which train and collect data using the same policy~\cite{Sutton2018reinforcement}. The two off-policy algorithms we test include Deep Q Networks (DQN)~\cite{Mnih2015DQN} and Advantage Actor Critic (A2C)~\cite{Mnih2016A2C}; the two on-policy algorithms we test include Trust Region Policy Optimization (TRPO)~\cite{Schulman2015TRPO} and Proximal Policy Optimization (PPO)~\cite{Schulman2017PPO}. 

\subsubsection{Comparison of Deep RL Algorithms}
\begin{figure}[!hbt]
\centering
\includegraphics[width=0.5\textwidth]{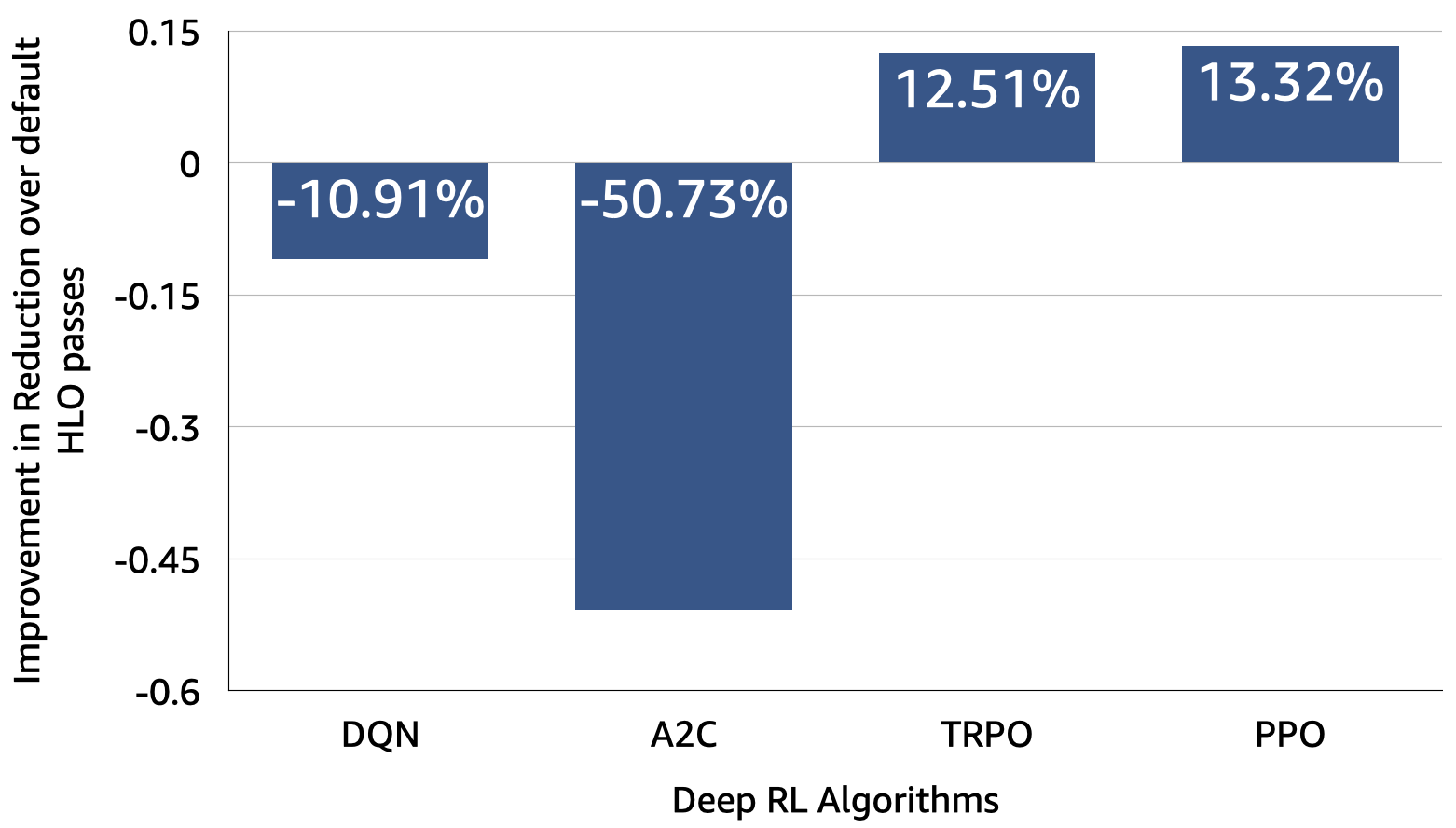}
\caption{Comparison of various RL algorithms.}
\label{fig:RL_ALG}
\end{figure}

We compare the results of benchmarking the various deep Reinforcement Learning algorithms shown in Figure~\ref{fig:RL_ALG}. The dataset we use for benchmarking is from a GPT-2~\cite{radford2019language} training loop implementation maintained internally in Amazon, adapted from the NVIDIA Megatron-LM~\cite{shoeybi2019megatron} project. The metric we use is the geometric mean over all the testing benchmarks of the ratio between the operation count reduction using the RL agent to that of the default HLO passes used in the AWS Neuron SDK~\cite{awsneuron2022}. Specifically: $\left(\prod _{b=1}^{B} \frac{(I_b - R_b)}{(I_b - D_b)} \right)^{\frac {1}{B}}$ where $I$ is initial XLA operation count, $R$ is count after using RL agent, $D$ is count using default HLO passes, $B$ is total number of benchmarks, and all counts are indexed by benchmark $b$. Overall, PPO performs the best with an average of $13.3\%$ improvement in XLA operation count reduction over the default HLO passes. However, the off-policy approaches had comparatively poor performance.  It is interesting to note that the work of~\cite{Huang2022a2cppo} shows A2C is equivalent to PPO when certain parameters are fixed — importantly the number of update epochs in PPO must be set to 1, there should be no clipping, and the KL divergence term in the loss is removed. Therefore we hypothesize these are what make PPO perform much better than A2C. Also, note the dataset used to train the RL model is different from (no intersection with) the dataset used to test the RL model and default HLO passes.

\begin{figure*}[t]
\centering
\includegraphics[width=\linewidth]{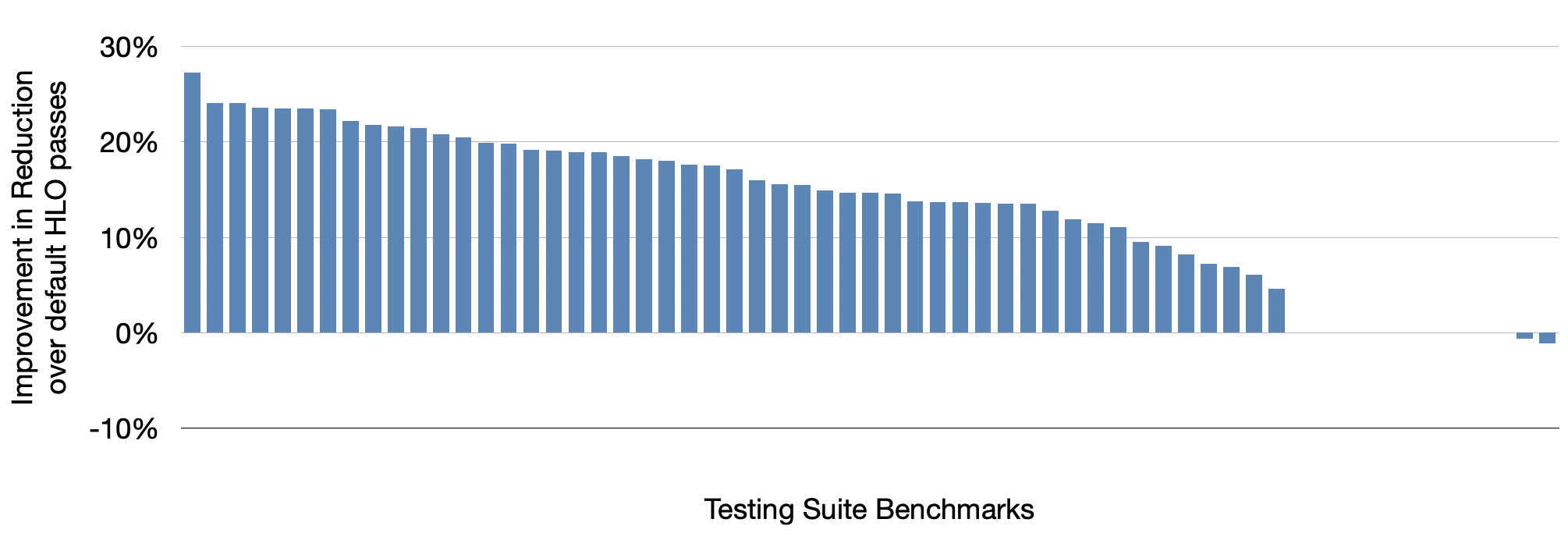}
\caption{We present the improvement in reduction using our proposed methodology over the default HLO passes for each benchmark in out testing suite. We are able to achieve up to $27.3\%$ total improvement. There are some benchmarks that have same performance as the default HLO as indicated by the gap on the right, likely because both approaches have reached near optimal optimization. On two benchmarks in our testing suite, we perform at most $1.2\%$ worse than the default HLO passes.}
\label{fig:Testing}
\end{figure*}

In Figure~\ref{fig:Testing} we provide the improvement in operation counts of our learning based approach over the default passes on various benchmarks. In around $97\%$ of the testing suite benchmarks, we can see that our methodology performs at least as good as the default HLO passes method. Furthermore, our approach is able to achieve up to $27.3\%$ total improvement in operation count reduction. 

\subsubsection{Execution Time Speedup}

So far we have been comparing using the HLO operation count as our metric for target-independent improvement. Using this metric, we are able to demonstrate that we can reduce our graph size in terms of the HLO operation counts using our proposed methodology. We now want to show how our proposed approach for improving XLA HLO optimization can translate to improving execution time without requiring any target-dependent optimizations. We utilize Tensorflow 2.9.2’s XLA CPU compiler to generate execution time computation on a m6g.16xlarge Amazon EC2 instance on the testing suite benchmarks. Ultimately, our methodology provides an average speedup of $1.0291\times$ with standard deviation of $0.0155$ (up to $1.0635\times$ speedup). 
The runtime speedup of $1.0291\times$ is somewhat expected because of decoupling, i.e., we are not using any target-specific cost function such as machine native instruction count for each XLA HLO operation.

\subsection{Evaluating Deep RL Algorithm Enhancements}
\label{section:EvaluateEnhancement}
Motivated by the improvement in performance provided by deep Reinforcement Learning algorithms, particularly PPO, we seek to further boost optimal policy search by introducing domain knowledge guidance. We explore how incorporating HLO cost analysis features in various aspects of the PPO algorithm affect performance. In particular, we examine our two proposed enhancements: reward shaping and value function modification.

\subsubsection{Reward Shaping}
Reward Shaping introduces an artificial reward signal to the environment feedback reward~\cite{Ng1999policy, Badnava2019}. This must come in the form of a potential function~\cite{Taylor2009}. Specifically, the work of~\cite{Ng1999policy} shows that function $F$ is a potential-based shaping function if there exists a real value function $\phi:\mathcal{S}\rightarrow \mathbb{R}$ so for all $s\in \mathcal{S} \setminus \{ s_0 \}$, $a\in \mathcal{A}$, and $s' \in \mathcal{S}$, then $F(s, s') = \gamma \phi(s') - \phi(s)$. Furthermore, for potential-based shaping function $F$, they prove that every optimal policy $\pi^*$ in MDP $M=\langle\mathcal{S},\mathcal{A},T,R,\gamma\rangle$ is also an optimal policy in MDP $M'=\langle\mathcal{S},\mathcal{A},T,R+F,\gamma\rangle$ and vice versa.

In this manner, we transform our initial MDP $M$ in XLA Gym to a new one $M'$ with provably same optimal policies by introducing a potential-based shaping function to the reward. We craft a heuristic based on HLO cost features like FLOP count and transcendental count into a function $\phi(s)$. Specifically $\phi(s)=-FLOP\_count(s) - 2*transcendental\_count(s)$. Therefore our new reward function will be $R'(s) = R(s) + \gamma*(-FLOP\_count(s') - 2*transcendental\_count(s')) - (-FLOP\_count(s) - 2*transcendental\_count(s) )= R(s) +\gamma\phi(s') - \phi(s)$.

\subsubsection{Value Function Modification}
Another approach we propose is to introduce the cost analysis features into the value function. In deep Reinforcement Learning algorithms, the value function captures the quality of the agent in a particular state. Specifically, it predicts the returns of the agent from that state. Cost analysis features may provide a better estimate of this quality. We introduce features like FLOP count and transcendental count to the value function so the value function takes the form $V(s, f)$ where $f$  is a vector of the additional cost analysis features.

\subsubsection{Comparison of Deep RL Algorithm Enhancements}
\begin{figure}[!hbt]
\centering
\includegraphics[width=0.5\textwidth]{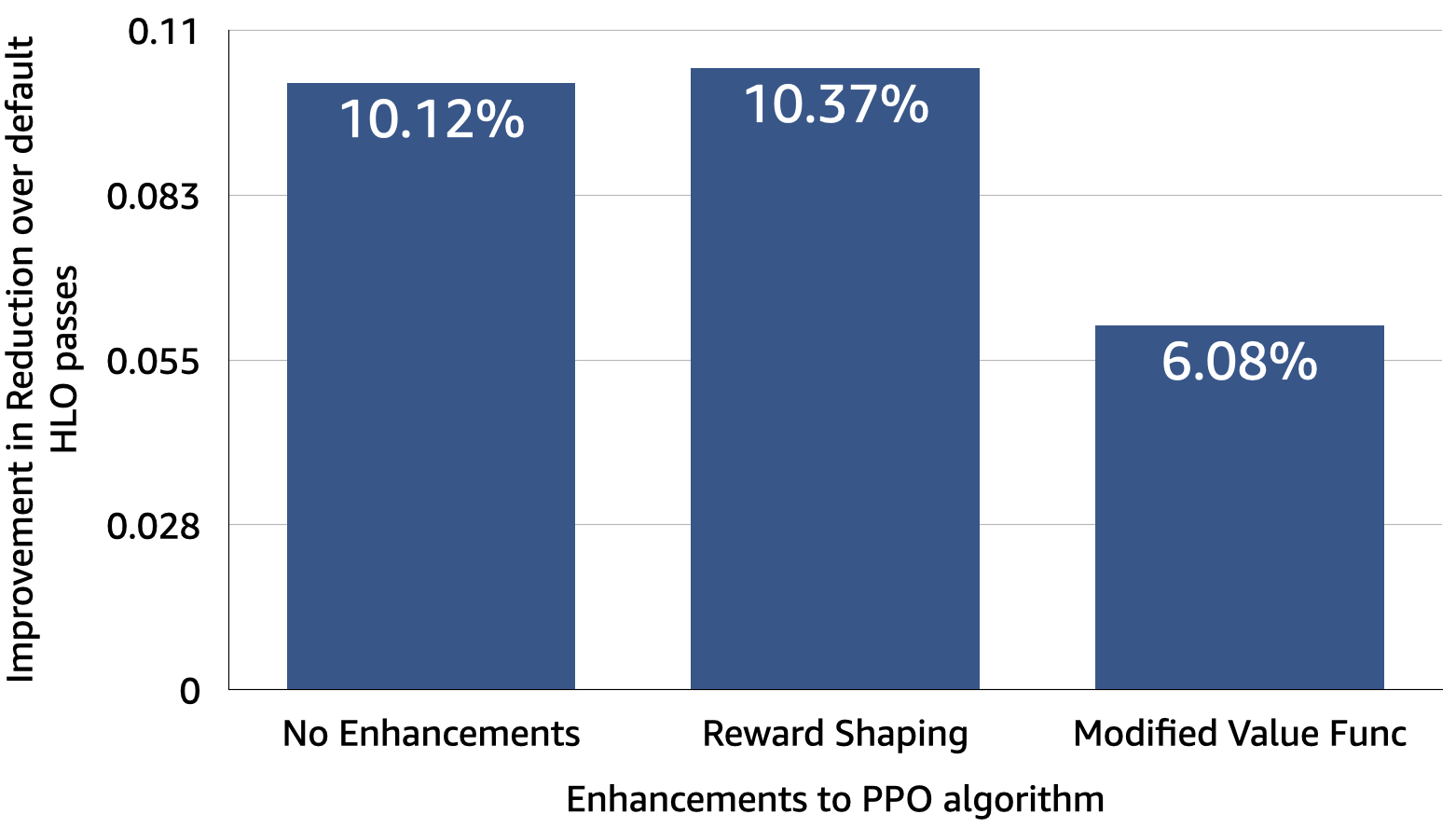}
\caption{Comparison of the results of enhancements to PPO.}
\label{fig:RL_ENHANCE}
\end{figure}

We accordingly test the enhancements to the PPO algorithm and compare it with the original PPO algorithm. The results can be seen in Figure~\ref{fig:RL_ENHANCE}. Note that in this comparison, we work with a much larger and more diverse data set of more than 300 graphs coming from models such as GPT-2, BERT~\cite{devlin2018bert}, ResNet~\cite{he2016deep} than from evaluation in Figure~\ref{fig:RL_ALG} for better comparison of robustness of the enhancements. Overall, PPO with the shaping potential does best in comparison to plain PPO. This demonstrates that there is more room for improvement in guiding the deep RL algorithms for optimal policy search by introducing domain specific knowledge. It is also interesting that the value function enhancement has poor performance. We hypothesize this may be due to what recent papers like~\cite{Chung2021} suggest that minimum variance baselines (Value function is baseline proxy used in PPO) do not necessarily correlate to convergence to optimal policy. In essence, although introducing cost analysis features may improve the value function's quality estimate accuracy, this may in the long run backfire by potentially encouraging the agent to commit and convergence to a suboptimal policy.

\section{Conclusion}
We have introduced a mostly unchartered problem of target-independent compiler optimization pass ordering in Machine Learning compilers like XLA. Specifically we propose reformulating the problem into a Markov Decision Process and address with Reinforcement Learning algorithms. We created an XLA Gym infrastructure with environments for XLA compiler optimization pass problem to specifically optimize XLA operation count, but this can be generalized to other optimization targets. Orthogonal extensions to our work include optimizing for execution speed and therefore exploring target-dependent methods like autotuning. Furthermore, we used an observation space of various types of XLA operation counts; however, there may be additional useful information in the graph structure itself. So, with much larger training suites to avoid overfitting, another future extension includes using a graph-based program representation like ProGraML~\cite{Cummins2021}. To test and demonstrate the effectiveness of using RL in HLO pass ordering, we tested various off-policy and on-policy deep Reinforcement Learning algorithms in XLA Gym. We also proposed and tested new enhancements to incorporate domain specific knowledge into the deep RL algorithms. Overall we achieve up to an average of $13.3\%$ improvement over the default HLO passes and show further room for improvement on deep RL algorithms through domain knowledge introduction.

\bibliographystyle{unsrt}

\newpage

\appendix

\section{Appendix}

\begin{table*}[h]
    \centering
    \begin{tabular}{c|c}
    Hyperparameters and Setup & Values \\\hline
    \textbf{On-policy parameters} &\\
    Batch size & $256$ \\
    Learning rate & $3e^{-4}$\\
    Discount factor & $0.99$\\
    Network Architecture & MLP $[2048,2048]$\\
    Entropy coefficient & 0 \\
    GAE lambda & 0.95 \\
    PPO clip range & 0.2 \\\hline
    \textbf{Off-policy parameters} &\\
    Batch size & $256$\\
    Buffer size & $10^6$\\
    Learning rate & $3e^{-4}$\\
    Discount factor & $0.99$\\
    Network Architecture & MLP $[2048,2048]$\\\hline
    \textbf{General parameters} &\\
    Training Steps & $200000$ \\
    Training time & average 11 hours\\
    Hardware & c5.18xlarge AWS EC2 Instance
    \end{tabular}
    \caption{Hyperparameter Settings Details}
    \label{appendix_table:parameter_settings}
\end{table*}

\end{document}